\documentclass[11pt, a4paper, logo, copyright]{googlecloud}

\pdftrailerid{redacted}

\makeatletter
\renewcommand\bibentry[1]{\nocite{#1}{\frenchspacing\@nameuse{BR@r@#1\@extra@b@citeb}}}
\makeatother

\usepackage[authoryear, sort&compress, round]{natbib}

\usepackage[utf8]{inputenc} % allow utf-8 input
\usepackage[T1]{fontenc}    % use 8-bit T1 fonts
\usepackage{hyperref}       % hyperlinks
\usepackage{url}            % simple URL typesetting
\usepackage{booktabs}       % professional-quality tables
\usepackage{amsfonts}       % blackboard math symbols
\usepackage{nicefrac}       % compact symbols for 1/2, etc.
\usepackage{microtype}      % microtypography
\usepackage{xcolor}         % colors

\usepackage[utf8]{inputenc} %
\usepackage[T1]{fontenc}    %
\usepackage{url}            %
\usepackage{booktabs}       %
\usepackage{nicefrac}       %
\usepackage{microtype}      %
\usepackage{amsmath}
\usepackage{graphicx}
\usepackage{multicol}
\usepackage{hyperref}       %
\usepackage[nameinlink]{cleveref}
\usepackage{bbm}
\usepackage{multirow}
\usepackage{soul}
\usepackage{floatrow}
\usepackage{float}
\usepackage{wrapfig}
\usepackage{blindtext}
\usepackage{tablefootnote}
\usepackage{amsfonts}
\usepackage[flushleft]{threeparttable}
\usepackage{bbding}
\usepackage{xcolor}
\usepackage{xspace}
\usepackage{bm}
\usepackage{arydshln}
\usepackage{subfigure}
\usepackage{enumitem}
\usepackage{setspace}
\usepackage{color}
\usepackage{algorithm}
\usepackage{longtable}

\usepackage[normalem]{ulem}
\usepackage{ulem}
\usepackage[nomargin,inline,marginclue,draft]{fixme}
\usepackage{balance}
\usepackage{verbatim}
\usepackage{diagbox}
\usepackage{changepage}
\usepackage{amssymb}
\usepackage{pifont}
\usepackage{array}   %

\usepackage{algpseudocode}

\usepackage{tcolorbox}

\usepackage{amsmath,amsfonts,bm}

\def\eqref#1{equation~\ref{#1}}

\def\1{\bm{1}}

\DeclareMathAlphabet{\mathsfit}{\encodingdefault}{\sfdefault}{m}{sl}
\SetMathAlphabet{\mathsfit}{bold}{\encodingdefault}{\sfdefault}{bx}{n}

\newcommand{\method}{\mbox{\textsc{DynScaling}}\xspace}

\title{\method: Efficient Verifier-free Inference Scaling via Dynamic and Integrated Sampling}

\correspondingauthor{fwang598@usc.edu, \{xingchenw,ruoxis,jiefengc,soarik\}@google.com}

\renewcommand{\today}{}

\author[1 2 *]{Fei Wang}
\author[1]{Xingchen Wan}
\author[1]{Ruoxi Sun}
\author[1]{Jiefeng Chen}
\author[1]{Sercan Ö. Arık}

\affil[1]{Google}
\affil[2]{University of Southern California}

\begin{abstract}
Inference-time scaling has proven effective in boosting large language model (LLM) performance through increased test-time computation. Yet, its practical application is often hindered by reliance on external verifiers or a lack of optimization for realistic computational constraints. We propose \method, which addresses these limitations through two primary innovations: an integrated parallel-sequential sampling strategy and a bandit-based dynamic budget allocation framework. The integrated sampling strategy unifies parallel and sequential sampling by constructing synthetic sequential reasoning chains from initially independent parallel responses, promoting diverse and coherent reasoning trajectories. The dynamic budget allocation framework formulates the allocation of computational resources as a multi-armed bandit problem, adaptively distributing the inference budget across queries based on the uncertainty of previously sampled responses, thereby maximizing computational efficiency. By combining these components, \method effectively improves LLM performance under practical resource constraints without the need for external verifiers. Experimental results demonstrate that \method consistently surpasses existing verifier-free inference scaling baselines in both task performance and computational cost.
\end{abstract}

\begin{document}

\maketitle

\section{Introduction}
Inference scaling boosts large language model (LLM) performance through increased test-time computation, offering a cost-effective alternative to retraining \citep{welleck2024decoding,snell2024scaling,chen2024more,wang2022self,beirami2024theoretical}. Early inference scaling efforts primarily aimed for peak performance using substantial computational budgets, frequently leveraging external verifiers like reward models \citep{brown2024large,wu2024inference,sardana2023beyond,li2022making}. Recognizing practical constraints, more recent approaches focus on maximizing performance within a fixed computational budget, acknowledging that this is often more critical for real-world users than achieving theoretical peaks with unlimited computation \citep{wang2025think,kang2025scalable,chen2025sets}. In addition, training and deploying verifiers \citep{cobbe2021training} incur extra cost, and in many practical scenarios, verifiers are either unavailable or unreliable \citep{gao2023scaling,liu2025can}. As a result, recent work has shifted toward more practical settings, emphasizing efficiency under \textit{manageable computational resources} and exploring \textit{verifier-free settings} to improve generalizability \citep{wang2025think,kang2025scalable,chen2025sets}.

Verifier-free\footnote{Using the LLM itself as the verifier is considered verifier-free.} inference scaling has emerged as a practical and effective strategy for enhancing the performance of LLMs. Recent work highlight that majority voting over responses obtained through a combination of sequential and parallel sampling remains among the most robust and widely applicable approaches across diverse models and tasks \citep{wang2025think,chen2025sets}. 
Despite the existence of multiple variants, these approaches generally follow a two-dimensional process: generating multiple independent responses in parallel, followed by iterative refinement of each response using the LLM itself in a sequential manner. However, the best practice for applying this strategy, including the proportion of parallel to sequential sampling and the design of the sequential refinement process, can vary across models, tasks, and even individual queries \citep{snell2024scaling}. These findings give rise to two central questions: (1) How can parallel and sequential inference be \textbf{effectively integrated} to optimize performance across diverse models and tasks?
(2) How can inference budgets be \textbf{dynamically allocated} throughout the sampling process based on intermediate results to improve overall efficiency?

To address the above problems, we propose \method, an efficient, verifier-free inference scaling approach tailored for practical deployment. Our method is grounded in two key innovations: an \textbf{integrated parallel–sequential sampling strategy} (\Cref{sec:method/integrated}) and a \textbf{bandit-inspired dynamic budget allocation framework} (\Cref{sec:method/dynamic}).
First, to better integrate the strengths 
of both parallel and sequential sampling, we design a strategy that treats the initially independent responses generated in parallel as building blocks for a synthetic sequential reasoning chain. Rather than sampling sequentially from scratch, \method constructs a chain by concatenating these parallel responses to motivate further thinking. To promote diversity in reasoning paths, we also apply random permutations when constructing the chain. This unified view allows \method to benefit from the breadth of parallel sampling and the depth of sequential refinement, while mitigating redundancy.
Second, to improve inference scaling efficiency, we formulate budget allocation as a multi-armed bandit (MAB) problem \citep{robbins1952some}. In this formulation, each query is considered as an arm, and the expected reward associated with pulling an arm corresponds to the uncertainty in the model’s current set of responses. Intuitively, queries with higher uncertainty are more likely to benefit from additional computation. We quantify this uncertainty using measures such as the inverse of the majority vote ratio, and adopt an Upper Confidence Bound (UCB) policy \citep{auer2002finite} to balance exploration and exploitation (\Cref{sec:method/ucb}). This dynamic allocation mechanism allows \method to focus computation where it is most needed, rather than uniformly distributing budget across all inputs.
Together, these components enable \method to achieve strong performance under manageable resources without relying on external verifiers.

We evaluate \method on two reasoning-intensive benchmarks, GPQA and AIME, using both lightweight non-thinking LLM and advanced thinking LLM. Compared to a range of representative verifier-free inference scaling methods, \method consistently achieves higher effectiveness, improved efficiency under practical budget constraints, and greater stability as the inference budget increases. These results underscore the strength of \method’s integrated and process-aware strategies. 

In summary, our contributions are as follows:
\begin{itemize}[left=5pt, labelsep=0.5em]
    \item We propose an \textbf{integrated parallel-sequential sampling strategy} that unifies parallel and sequential sampling by synthesizing multiple independently generated responses into diverse and coherent reasoning trajectories.
    \item We introduce a \textbf{dynamic budget allocation framework} formulated as a multi-armed bandit problem, which adaptively adjusts the inference budget across queries based on the uncertainty of previously sampled responses, thereby improving computational efficiency.
    \item By combining the above components, we present \method, an \textbf{efficient verifier-free inference scaling method}. We empirically demonstrate that \method consistently outperforms representative verifier-free inference scaling baselines in terms of effectiveness, efficiency, and stability.
\end{itemize}

\section{Related Work}

\subsection{Inference Scaling for LLMs}
Inference scaling has emerged as a practical strategy to boost the performance of LLMs at test time without requiring retraining. Early work in this area focused on maximizing performance by leveraging extensive computational budgets, often incorporating external verifiers, such as reward models, to guide fine-grained tree search \citep{feng2023alphazero,wu2024inference} or identify the best response \citep{brown2024large}. While these methods can be effective, their focus on high computational cost regime and dependence on expensive or unreliable verifiers limit their practicality for real-world applications.
Recent research has shifted toward more realistic settings that prioritize efficiency under constrained computational budgets, while avoiding the cost and complexity of training or deploying additional verifiers. Following this direction, \citet{wang2025think} benchmark a series of existing methods and identify majority voting as one of the most robust verifier-free strategies. \citet{chen2025sets} leverage LLMs themselves as implicit verifiers to develop a structured sequential sampling framework. Other approaches, such as \citet{kang2025scalable}, assume access to token-level logits to develop fine-grained metric to guide generation. Our work builds on this line of research with minimal LLM access by proposing a verifier-free, model-agnostic method that integrates the strengths of sequential and parallel inference, while dynamically allocating budget across queries to balance effectiveness and efficiency.

\subsection{Adaptive Inference Budget Allocation for LLMs}
Recent work has explored how to adaptively allocate inference budget either within or across queries for LLMs. Some approaches focus on intra-query adaptation, adjusting internal generation strategies rather than distributing a fixed budget across multiple queries, and are thus orthogonal to our setting. For example, Osca \citep{zhang2024scaling} dynamically adjusts the sampling budget for different inference configurations within a query, while DISC \citep{light2025disc} uses a binary decision process to refine selected parts of a previous response. MetaScale \citep{liu2025metascale} employs a UCB-style method to select prompts with diverse personas, enhancing response diversity.
In contrast, our work focuses on inter-query budget allocation. Earlier efforts in this direction began with simplified setups, such as those tailored solely for self-consistency \citep{aggarwal2023let}. The most relevant prior efforts include \citet{damani2024learning}, who train a separate classifier to predict query difficulty based on supervision from a reward model; \citet{manvi2024adaptive}, who fine-tune the LLM to act as a generative reward model that determines whether further sampling is needed; and \citet{yu2025think}, who learn to control the distribution of response lengths under a constrained reinforcement learning framework. However, all of these methods require additional training and rely on external supervision. Our approach is fully inference-time, model-agnostic, and leverages uncertainty signals from the model's own outputs without any fine-tuning or auxiliary models.

\begin{algorithm}[t]
\caption{Bandit-based Dynamic Budget Allocation}
\label{alg:adaptive}
\begin{algorithmic}[1]
\Require Query set $Q = \{q_1, \dots, q_m\}$, LLM $\mathcal{M}$, Total budget $B_{\text{total}}$, Unit budget $B_{\text{unit}}$
\State Initialize $B_{\text{used}} \gets 0$
\For{$q_i \in Q$} \Comment{\textcolor{blue}{Sample initial responses for each query}}
    \State $R_i \gets \texttt{integrated\_sampling}(q_i, \mathcal{M}, B_{\text{unit}})$  \Comment{\textcolor{blue}{\Cref{alg:integrated_sampling}}}
    \State $B_i \gets B_{\text{unit}}$
    \State $B_{\text{used}} \gets B_{\text{used}} + B_{\text{unit}}$
\EndFor
\While{$B_{\text{used}} < B_{\text{total}}$} \Comment{\textcolor{blue}{Allocate more samples dynamically for prioritized queries}}
    \State Compute $a_i \gets \texttt{sampling\_priority}(R_i, B_i, B_{\text{used}})$ for each $q_i$  \Comment{\textcolor{blue}{\Cref{alg:sampling_priority}}}
    \State Select subset $\mathcal{S} \subseteq Q$ with top-ranked $a_i$
    \For{$q_i \in \mathcal{S}$}
        \State $R_i \gets R_i \cup \texttt{integrated\_sampling}(q_i, \mathcal{M}, B_{\text{unit}})$  \Comment{\textcolor{blue}{\Cref{alg:integrated_sampling}}}
        \State $B_i \gets B_i + B_{\text{unit}}$
        \State $B_{\text{used}} \gets B_{\text{used}} + B_{\text{unit}}$
    \EndFor
\EndWhile
\For{$q_i \in Q$} \Comment{\textcolor{blue}{Majority voting for the final answer}}
    \State $\hat{r}_i \gets \texttt{majority\_vote}(R_i)$
\EndFor
\State \Return Final responses $\{\hat{r}_i\}_{i=1}^m$
\end{algorithmic}
\end{algorithm}

\section{Methods}  
\label{sec:method}
We introduce \method, a verifier-free inference scaling approach for LLMs that achieves more judicious utilization of a fixed inference budget by simultaneously employing parallel and sequential sampling and dynamically allocating inference resources across queries.
\method integrates the strengths of parallel and sequential sampling by linking initially independent responses into synthetic thought chains, combining the diversity of parallel sampling with the reasoning depth of sequential prompting. This fusion enhances answer quality at a reduced cost. To further improve efficiency, \method employs a bandit-based dynamic allocation mechanism that adjusts per-query budgets on the fly, guided by signals from earlier responses that balance exploitation and exploration. We begin with a high-level overview of the framework, followed by a detailed description of its core algorithmic components.

\subsection{Overview}
\label{sec:method_overview}
Given a fixed total inference budget, \method aims to maximize LLM output quality across a batch of queries by optimizing how responses are sampled and where the budget is spent. It consists of two key ingredients: (1) an integrated strategy that combines parallel and sequential sampling, and (2) a multi-armed bandit (MAB)-based approach to dynamically prioritize more challenging queries where the model faces uncertainty, so that they can benefit from additional computation.
Specifically, \method begins by generating a small set of diverse responses per query using parallel sampling. These are then composed into synthetic thought chains and used as conditional context for additional samples, incorporating sequential reasoning without committing to fixed-depth chains. This integrated sampling approach produces more informative responses with limited cost. Meanwhile, an uncertainty-driven priority mechanism identifies which queries are challenging for the model, dynamically allocating additional sampling budget through a Upper Confidence Bound (UCB)-style acquisition strategy.

\begin{algorithm}[t]
\caption{Integrated Parallel-Sequential Sampling}
\label{alg:integrated_sampling}
\begin{algorithmic}[1]
\Require Query $q$, Model $\mathcal{M}$, Budget $B$, Thought length $k$
\State $R_{\text{init}} \gets \texttt{sample}(\mathcal{M}, q, B/2)$ \Comment{\textcolor{blue}{Generate initial responses in parallel}}
\State $R_{\text{cond}} \gets [\ ]$
\For{$j = 1$ to $B/2$}
    \State $T_j \gets \texttt{concat}(\texttt{random\_sample}(R_{\text{init}}, k))$ \Comment{\textcolor{blue}{Combine parallel samples as a sequence}}
    \State $r_j \gets \texttt{sample}(\mathcal{M}, q + T_j, 1)$ \Comment{\textcolor{blue}{Sample more responses based on the integrated sequence}}
    \State Append $r_j$ to $R_{\text{cond}}$
\EndFor
\State \Return $R_{\text{init}} \cup R_{\text{cond}}$
\end{algorithmic}
\end{algorithm}

\begin{algorithm}[t]
\caption{UCB-based Sampling Priority}
\label{alg:sampling_priority}
\begin{algorithmic}[1]
\Require Responses $R_i$, Budget $B_i$, Total used budget $B_{\text{used}}$, Exploration ratio $c$
\State $u_i \gets \texttt{uncertainty}(R_i)$ \Comment{\textcolor{blue}{Compute uncertainty as reward}}
\State $a_i \gets u_i + c \cdot \sqrt{\frac{\log(B_{\text{used}})}{B_i}}$ \Comment{\textcolor{blue}{Upper Confidence Bound (UCB)}}
\State \Return $a_i$
\end{algorithmic}
\end{algorithm}

\subsection{Integrated Parallel-Sequential Sampling}
\label{sec:method/integrated}
Traditional sampling-based inference methods separately extend parallel sampling (which provides diversity through independent generation) and sequential prompting (which ensures depth through chain-of-thought reasoning). 
In parallel sampling, the sequences are independent and do not benefit from each other, while sequential methods require multiple steps to reach a fixed depth, making them inefficient. This separation, where parallel sampling focuses on diversity and sequential sampling focuses on depth, often results in suboptimal performance, especially when working within a fixed budget. 
To effectively leverage the strengths of both strategies, we propose Integrated Parallel-Sequential Sampling (\Cref{alg:integrated_sampling}). Given a per-query budget $B$, we begin by generating $B/2$ independent completions through parallel sampling. Next, we synthesize these responses into intermediate "thought segments" by randomly selecting and concatenating them to create prompt continuations. These synthetic chains provide soft guidance for the remaining $B/2$ samples, encouraging the model to refine or extend promising reasoning from the initial outputs. This hybrid sampling protocol strikes a balance between diversity and depth, enhancing answer quality without requiring costly multi-turn prompting. By integrating parallel and sequential sampling, our approach allows the two to complement and enhance each other, resulting in a more efficient allocation of resources.

\subsection{MAB-Based Dynamic Budget Allocation}
\label{sec:method/dynamic}

To further improve budget efficiency across a batch of queries, we introduce a dynamic allocation mechanism (\Cref{alg:adaptive}) that adaptively decides which queries should receive higher sampling budget. Inspired by the MAB framework, our method treats each query as a bandit arm and distributes a fixed overall budget by iteratively allocating unit increments to the most promising queries.
The process begins with an initial allocation of the unit budget $B_{\text{unit}}$ to each query, used to generate a small set of responses via Integrated Parallel-Sequential Sampling. As more samples are collected, the model tracks a priority score for each query, which reflects how informative or uncertain the current responses are. At each step, a subset of high-priority queries is selected to receive additional budget. This allocation continues until the total budget $B_{\text{total}}$ is exhausted.
This dynamic strategy enables the model to utilize its generation budget more wisely by investing additional resources in queries that are uncertain or potentially more difficult based on model-specific signals, while avoiding over-sampling on queries for which additional sampling is unlikely to change the final predicted answer. As a result, \method balances coverage across all inputs while adapting sampling effort according to model-perceived difficulty.

\subsection{UCB-Style Process-Aware Sampling Priority}
\label{sec:method/ucb}

Our intuition is that queries eliciting higher uncertainty from an LLM are more likely to benefit from additional sampling. In such cases, generating more responses can increase the chance of uncovering a correct or high-quality answer. Therefore, we use the model's own uncertainty captured from its generated responses as a signal for reward, guiding where to allocate more budget.
To operationalize this idea, we adopt a variant of the UCB algorithm (\Cref{alg:sampling_priority}), commonly used in MABs. In our setting, each query acts as an arm, and the model decides which queries should receive more sampling budget based on both their observed uncertainty and how much they have already been sampled. Specifically, we assign each query $q_i$ a priority score:
\[
a_i = u_i + c \cdot \sqrt{\frac{\log B_{\text{used}}}{B_i}},
\]
where:
\begin{itemize}
    \item $u_i$ is the reward term representing uncertainty over the current responses for $q_i$,
    \item $B_i$ is the total budget spent so far on query $q_i$,
    \item $B_{\text{used}}$ is the total budget consumed across all queries,
    \item $c$ is a tunable hyperparameter that controls the trade-off between exploration and exploitation.
\end{itemize}

The first term, $u_i$, encourages \textit{exploitation} by prioritizing queries where the model exhibits high uncertainty and thus stands to gain from further sampling. The second term promotes \textit{exploration} by favoring queries that have been sampled less frequently, ensuring a fair chance of improvement across all queries.
We define \( u_i \) as the proportion of non-majority predictions (i.e., variation ratio) among current samples:
\[
u_i = 1 - \max_{a \in \mathcal{A}_i} \frac{\text{count}(a)}{n_i},
\]
where \( \mathcal{A}_i \) is the set of unique answers sampled for query \( q_i \), and \( n_i \) is the number of samples. This measure reflects the model’s uncertainty: it is high when responses are diverse and low when most responses agree. Intuitively, higher \( u_i \) indicates that the model has not converged on a dominant answer and may benefit from further sampling.
This process-aware priority mechanism adapts to the model’s evolving generation behavior. Rather than relying on static notions of difficulty, it tailors budget allocation to the model's own uncertainty signals. As a result, our approach focuses resources on queries that are challenging from the model's perspective, leading to a more effective and efficient use of the inference budget.

\section{Experiments}
\label{sec:experiment}
We first describe the experimental setup (\Cref{sec:experiment_setting}) and then present the main results, comparing \method with verifier-free inference scaling baselines for LLMs under varying budget levels (\Cref{sec:experiment_results}).
Finally, we provide detailed analyses on different modules (\Cref{sec:experiment_analysis}).

\subsection{Experimental Settings}
\label{sec:experiment_setting}

\paragraph{Datasets.} 
We conduct experiments on three datasets widely used to evaluate LLMs, spanning diverse domains and reasoning challenges.
\textbf{GPQA Diamond} \citep{rein2024gpqa} is a highly challenging subset of the GPQA benchmark, comprising 198 multiple-choice questions that require PhD-level knowledge in the natural sciences, including biology, chemistry, and physics. Throughout this paper, we refer to this subset simply as GPQA.
\textbf{AIME 24\footnote{\url{https://huggingface.co/datasets/HuggingFaceH4/aime_2024}} and AIME 25\footnote{\url{https://huggingface.co/datasets/opencompass/AIME2025}}} are math reasoning datasets derived from the 2024 and 2025 American Invitational Mathematics Examinations, respectively. Each dataset contains 30 problems that test advanced mathematical problem-solving skills, often involving nontrivial, multi-step reasoning. For evaluation, we combine the two sets into a single benchmark. Throughout this paper, we refer to this combined benchmark as AIME.

\paragraph{LLMs.}
We conduct experiments using two widely adopted LLMs, with and without the thinking mode introduced by \citet{guo2025deepseek}.
\textbf{Gemini 1.5 Flash} \citep{team2024gemini} is a lightweight variant from the Gemini 1.5 family, optimized for efficiency with minimal degradation in quality. Due to its low cost and latency, it is commonly used in inference-scaling scenarios. This model does not support thinking mode and typically produces concise responses. Unless otherwise specified, we set the temperature to 0.6 and the maximum output length to 8,192 tokens.
\textbf{Gemini 2.5 Pro}\footnote{\url{https://blog.google/technology/google-deepmind/gemini-model-thinking-updates-march-2025/}} is among the most advanced LLMs, achieving state-of-the-art performance across a wide range of tasks. It supports thinking mode \cite{guo2025deepseek,qwen3}, which enables a structured internal reasoning process before finalizing a response. By default, we enable thinking mode for all responses and set the temperature to 0.6 and the maximum output length to 32,768 tokens.

\paragraph{Metric.}
Following prior work \citep{wang2025think,chen2025sets,muennighoff2025s1}, we extract the model's predicted answer using pattern matching techniques and evaluate its correctness based on strict equivalence with the ground-truth answer. 
To assess how performance varies with computational resources, we report model accuracy as a function of the inference budget. However, due to inherent variability, such as changes in sampling depth and breadth when fully utilizing the allocated budget, model performance can fluctuate sharply across nearby budget levels.
To reduce this noise and reveal more stable trends, we apply a moving average smoothing technique. Specifically, the reported accuracy at each budget point is computed as the average over a sliding window of adjacent budget values. This helps present a clearer picture of overall performance without being overly sensitive to local variations.
All reported results are averaged over three independent runs to account for stochasticity in model generation.

\paragraph{Baselines.}
We compare \method against a range of \textit{verifier-free inference baselines}, which aim to enhance model performance without relying on external verifiers. Broadly, inference scaling strategies can be categorized along two axes: \textit{sampling breadth}, which involves generating multiple responses in parallel, and \textit{sampling depth}, which expands the reasoning process sequentially. Building on the taxonomy introduced by \citet{wang2025think}, we group the baselines into four categories based on how they implement sequential budget expansion:

\begin{itemize}
    \item \textbf{BoN (Majority)}: This baseline performs \textit{majority voting} over multiple independently sampled responses. It does not involve any sequential expansion. Despite its simplicity, it remains one of the most effective verifier-free inference scaling strategies, as recognized by \citet{wang2025think}.
    
    \item \textbf{SP1}: This approach involves a \textit{single-step} sequential refinement. Typically, a trigger word or phrase (e.g., ``Wait'') is appended to the initial response to prompt the model to continue refining its answer, such as \textit{budget forcing} \citep{muennighoff2025s1}.
    
    \item \textbf{SP2}: This class of methods performs sequential expansion in \textit{two stages}. First, the model generates feedback on its previous answer; second, it refines the previous response based on that feedback \citep{li2024hindsight}. Representative examples include the \textit{self-verification and correction (SETS)} framework proposed by \citet{chen2025sets}. In our experiments, we adopt the prompt variant from \citet{wang2025think}.
    
    \item \textbf{SP3}: This approach extends SP2 by incorporating a \textit{third step}: the model evaluates its responses and assigns each a score. Instead of using simple majority voting, the final answer is selected based on a voting mechanism that incorporates these self-assigned scores. We adopt the \textit{combined sequential parallel} framework introduced by \citet{wang2025think} for implementing SP3.
\end{itemize}

\paragraph{Implementation Details.}  
We use the number of samples as the budget unit, as it is easier to control, particularly when access to LLMs is limited. Specifically, we set the budget unit \( B_{\text{unit}} \) to 8 samples. This choice ensures that our integrated parallel-sequential sampling yields \( \frac{B_{\text{unit}}}{2} = 4 \) predicted answers, which we find to be the minimal effective number for majority voting. The thought length \( k \) is set to 4, as increasing \( k \) further provides negligible performance gains in our observations. We fix the exploration ratio \( c \) to \( \frac{1}{4} \) by default, and analyze its impact in \Cref{sec:experiment_analysis}.

\subsection{Main Results}
\label{sec:experiment_results}

\begin{figure}
    \centering
    \includegraphics[width=1\linewidth]{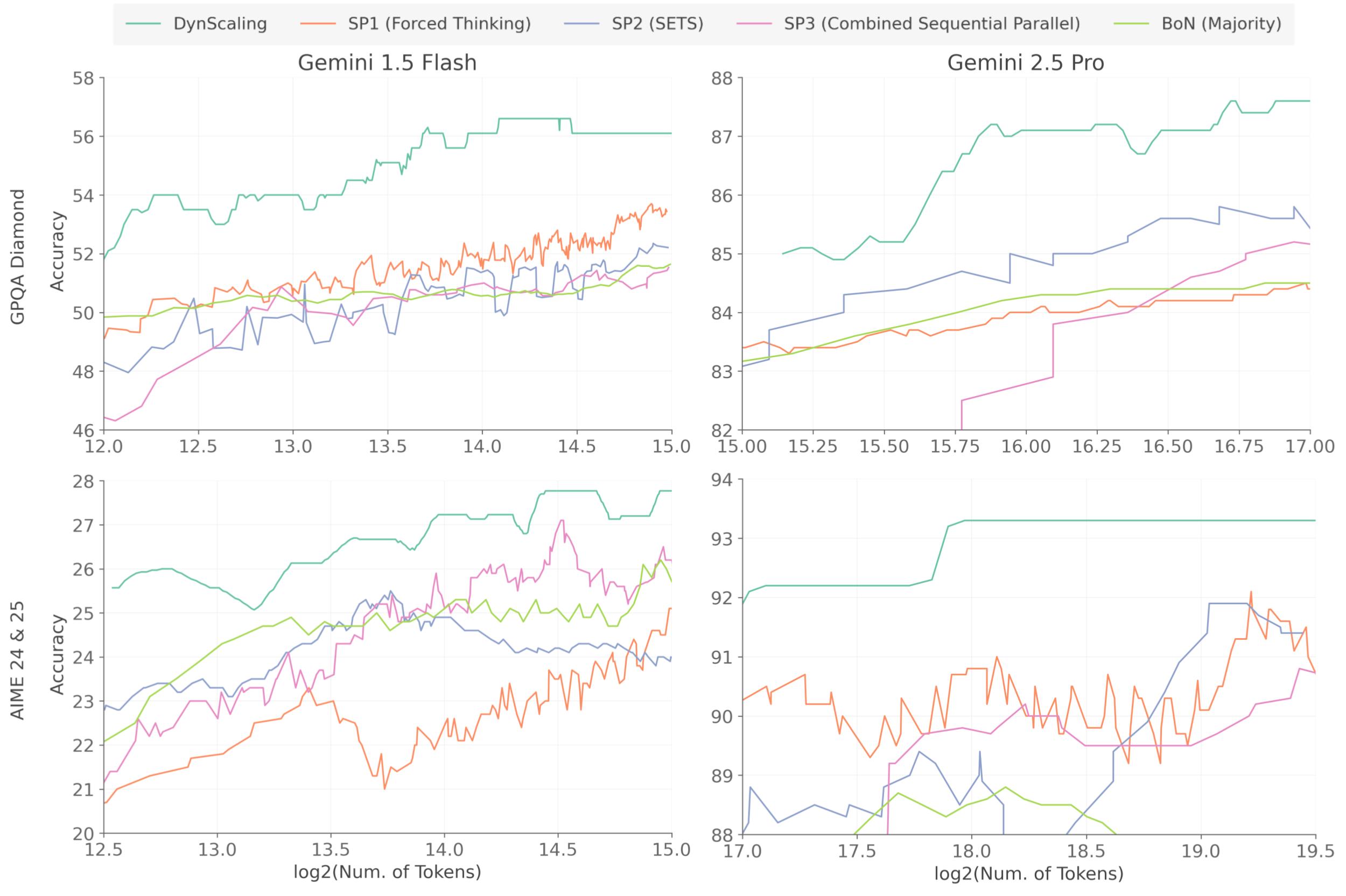}
    \caption{Performance of Gemini 1.5 Flash and Gemini 2.5 Pro on GPQA and AIME across varying inference budgets (measured by the number of output tokens). \method consistently outperforms representative verifier-free inference scaling baselines, demonstrating greater effectiveness, efficiency, and stability across budgets.}
    \label{fig:main_result}
\end{figure}

\paragraph{Overall Superiority of \method.}
As shown in \Cref{fig:main_result}, \method consistently outperforms all verifier-free inference scaling baselines across settings, offering superior effectiveness (achieving better performance at the same cost), efficiency (achieving the same performance at a lower cost), and stability (exhibiting smoother trends as the budget increases). These advantages hold across both lightweight models (Gemini 1.5 Flash) and advanced thinking models (Gemini 2.5 Pro), and span diverse tasks including scientific (GPQA) and mathematical (AIME) reasoning.

\paragraph{Efficiency in Low-Budget Regimes.}
\method demonstrates particularly strong gains under limited inference budgets. For instance, on GPQA with Gemini 1.5 Flash (top-left), it consistently outperforms the best baseline by around 3 accuracy points across most budget levels while maintaining a smoother growth curve. On AIME (bottom-left), it similarly outperforms other methods throughout the budget spectrum, showcasing superior sample efficiency. These results underscore the effectiveness of our dynamic budget allocation strategy under practical, resource-constrained inference budgets.

\paragraph{Scaling Behavior with Thinking Models.}
With the more capable Gemini 2.5 Pro, \method continues to outperform baselines. On GPQA (top-right), it scales effectively with budget while maintaining a consistent and sizable margin over other approaches. On AIME (bottom-right), although all methods show increased variance at higher budgets, \method avoids the noisy performance drops observed for SP2 and SP3, remaining among the top-performing methods throughout.

\paragraph{Comparison between Sequential and Parallel Strategies.}
BoN, which scales inference breadth without any sequential refinement, provides steady but modest gains. SP1–SP3 incorporate increasingly complex (structured) sequential prompting but exhibit unstable performance, often fluctuating across budget levels and failing to match the robustness of \method. This highlights the limitations of treating parallel and sequential strategies in isolation, and the advantage of \method's integrated parallel-sequential sampling approach.

\begin{figure}[t]
    \centering
    \includegraphics[width=\linewidth]{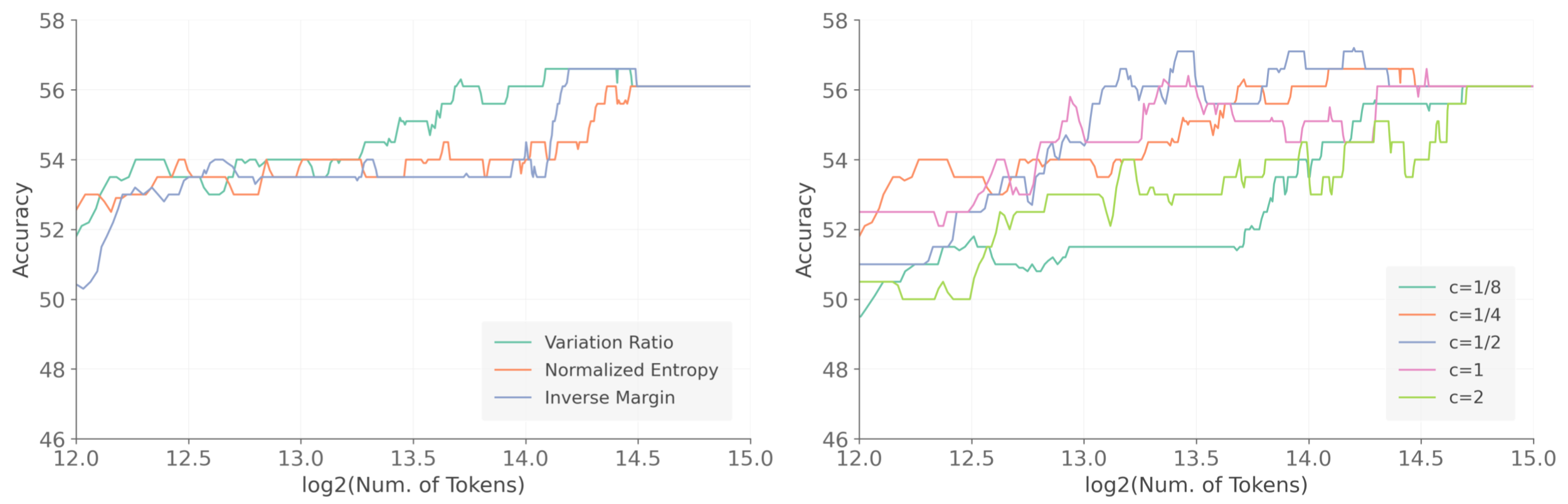}
    \caption{Analysis using Gemini 1.5 Flash on GPQA. \textbf{Left:} Comparison of different uncertainty measurements. \textbf{Right:} Effect of varying the exploration ratio.}
    \label{fig:analysis}
\end{figure}

\subsection{Analyses}
\label{sec:experiment_analysis}
We examine how different uncertainty measurements and exploration ratios affect performance as a function of the inference budget.
We conduct analyses using the Gemini 1.5 Flash model on the GPQA dataset. Unless otherwise specified, all experiments are run with default settings. 

\paragraph{Comparison of Different Uncertainty Measurements}
As shown in \Cref{fig:analysis} (left), all three uncertainty estimation methods, including Variation Ratio, Normalized Entropy, and Inverse Margin, exhibit a consistent trend where accuracy improves as more budget is processed, eventually converging to similar values. These results suggest that \method is robust to the choice of uncertainty estimation.
Among them, \textit{Variation Ratio} shows a more consistent and gradual increase in accuracy across the entire token range, maintaining competitive performance throughout. \textit{Normalized Entropy} provides steady accuracy in the lower budget range, followed by an increase at high budget range that brings it on par with Variation Ratio. \textit{Inverse Margin}, while initially underperforming, displays two distinct surges in accuracy at specific budget intervals, eventually catching up with the other methods at higher budget.

\begin{figure}[t]
    \centering
    \includegraphics[width=\linewidth]{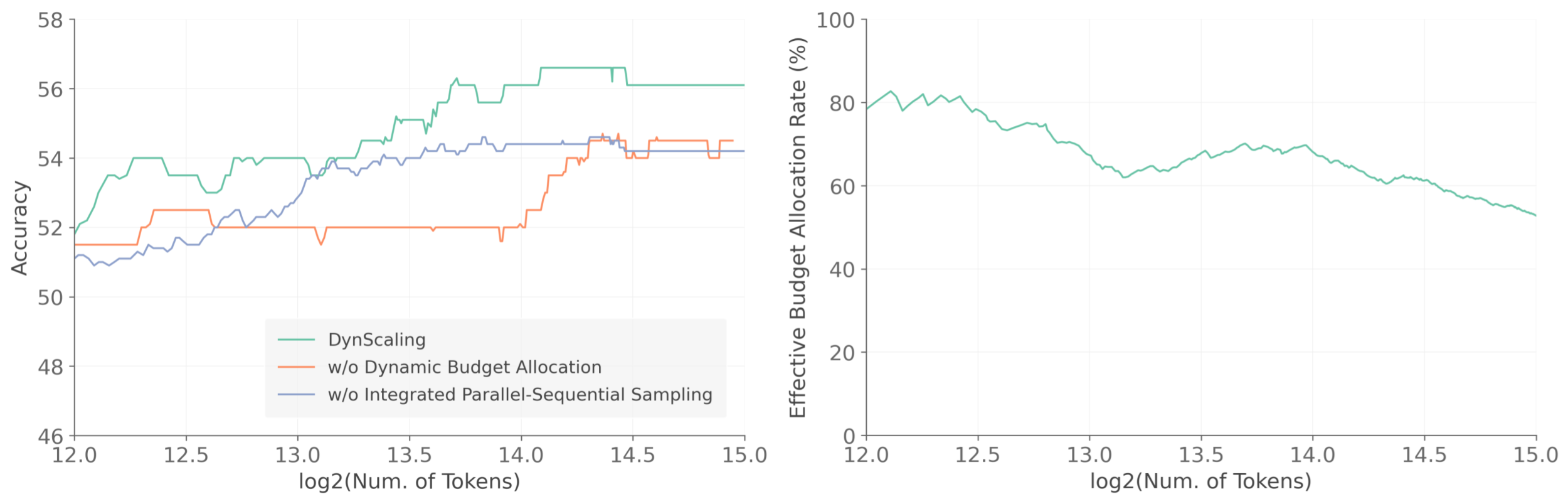}
    \caption{Analysis using Gemini 1.5 Flash on GPQA. \textbf{Left:} Ablation study of \method components. \textbf{Right:} Effective budget allocation toward incorrect queries.}
    \label{fig:analysis_2}
\end{figure}

\paragraph{Effect of Different Exploration Ratios.}
We evaluate the impact of varying the exploration ratio $c$ across values of $\frac{1}{8}$, $\frac{1}{4}$, $\frac{1}{2}$, $1$, and $2$, observing consistent trends in accuracy progression in \Cref{fig:analysis} (right). Smaller ratios, particularly $c = \frac{1}{8}$ and $c = \frac{1}{4}$, tend to yield higher initial accuracy under limited budget. In contrast, larger values such as $c = \frac{1}{2}$, $c = 1$, and $c = 2$ often catch up or surpass the smaller ratios as the budget increases, benefiting from stronger exploratory behavior in longer processing regimes. While no single value of $c$ is universally optimal, the results suggest that smaller ratios are better suited for low-budget settings, whereas larger ratios offer advantages when more budget is available. In practice, the choice of $c$ can be guided by the expected inference budget.

\paragraph{Ablation Studies of \method Components} 
We conduct an ablation analyses to isolate the contributions of key components in \method. As shown \Cref{fig:analysis_2} (left), removing either \textit{Dynamic Budget Allocation} or \textit{Integrated Parallel-Sequential Sampling} leads to noticeable drops in performance across most budget levels. The complete method consistently outperforms both ablated versions, particularly in the early to mid stages, demonstrating the benefits of effective early exploration and adaptive allocation. The sharp plateau of the \textit{w/o Dynamic Budget Allocation} variant highlights the rigidity introduced by a static budget strategy, while the slower early growth of the \textit{w/o Integrated Parallel-Sequential Sampling} variant reflects the limitations of decoupled sequential and parallel sampling. These trends confirm that both components are essential for the efficiency and accuracy gains achieved by \method, especially under tight or moderate budget constraints.

\paragraph{Effective Budget Allocation Toward Incorrect Queries} 
We analyze the \textit{effective budget allocation rate}, defined as the proportion of additional budget allocated to the remaining incorrect queries. As shown in \Cref{fig:analysis_2} (right), \method maintains a consistently high allocation rate across varying budget levels. When the total budget is low, many incorrect queries are relatively easy to detect, allowing \method to allocate up to 80\% of the additional budget toward correcting them. As the budget increases, fewer incorrect queries remain, and those that do are harder to identify, leading to a gradual decline in the allocation rate. Nevertheless, even in the most challenging cases within the studied range, the effective allocation rate remains above 50\%, highlighting \method's robustness in prioritizing error correction even under diminishing returns.

\section{Conclusion}

We presented \method, a practical and verifier-free inference scaling approach that integrates the strengths of both parallel and sequential sampling while dynamically allocating inference budgets. By synthesizing independent responses into a unified reasoning chain and formulating budget allocation as a bandit problem, \method effectively balances computational cost and performance. Our experiments on challenging reasoning benchmarks demonstrate that \method achieves superior effectiveness, efficiency, and robustness compared to existing verifier-free baselines. These results highlight the potential of process-aware and uncertainty-guided inference strategies for practical deployment of LLMs under constrained resources.

\section*{Limitations} 
While \method is model-agnostic and operates entirely at inference time, it assumes access to a batch of queries to enable cross-query budget reallocation. This requirement may restrict its applicability in latency-sensitive or single-query scenarios, such as real-time interactive systems. Moreover, \method currently relies on simple uncertainty heuristics, which may not capture nuanced ambiguity in model responses and could limit allocation precision in more complex settings. Furthermore, although we focus on reasoning benchmarks in this work, the generalizability of \method to other domains or task types remains to be explored.

\section*{Future Work} 
To further broaden the scope of \method, future work can (1) extend our approach to support fine-grained, token-level or intra-query budget allocation for more nuanced control over inference computation; (2) incorporate richer signals, such as alignment with task-specific priors, to refine the exploitation aspect of the budget allocator; and (3) adapt \method for streaming or interactive settings, where queries arrive incrementally and immediate, per-query decisions are required without access to a query batch.

\section*{Broader Impacts}
Our work on \method enhances the efficiency and practicality of LLM inference, potentially broadening access to advanced AI capabilities in domains such as education, healthcare, and scientific research. By eliminating the need for external verifiers, \method lowers deployment barriers and reduces computational costs, making powerful LLMs more accessible to a wider range of users and applications.
While \method itself does not introduce new ethical risks, careful evaluation and responsible use remain essential. Future work should explore safeguards and monitoring mechanisms to mitigate potential negative societal impacts.

\section*{Acknowledgement}

We would like to thank Jinsung Yoon for valuable discussions and insights that helped to improve this paper. 
We would also like to thank all other colleagues from Google for their valuable feedback.

{
\small
\bibliographystyle{unsrtnat}
\bibliography{reference}
}

\end{document}